\documentclass[journal]{IEEEtran}
\usepackage{amsmath,amsfonts}
\usepackage{algorithm}
\usepackage{algpseudocode}
\usepackage{array}
\usepackage[caption=false,font=normalsize,labelfont=sf,textfont=sf]{subfig}
\usepackage{textcomp}
\usepackage{stfloats}
\usepackage{url}
\usepackage{verbatim}
\usepackage{graphicx}
\usepackage{cite}
\usepackage{siunitx}
\usepackage{multirow}

\begin{document}

\title{Data-efficient Tactile Sensing with Electrical Impedance Tomography}

\author{Huazhi Dong,\IEEEmembership{ Student Member, IEEE},
Ronald B. Liu,\IEEEmembership{ Student Member, IEEE},
Leo Micklem,
Peisan (Sharel) E,\IEEEmembership{ Member, IEEE}, 
Francesco Giorgio-Serchi,\IEEEmembership{ Member, IEEE}, 
and Yunjie Yang,\IEEEmembership{ Senior Member, IEEE}
        % <-this % stops a space
\thanks{This work was supported in part by the European Research Council Starting Grant under Grant no.101165927 (Project SELECT).}% <-this % stops a space
\thanks{Manuscript received April 19, 2021; revised August 16, 2021.(Corresponding author: Yunjie Yang.)}

\thanks{Huazhi Dong and Yunjie Yang are with the SMART Group, Institute for Imaging, Data and Communications, School of Engineering, The University of Edinburgh, EH9 3BF Edinburgh, U.K. (e-mail: huazhi.dong@ed.ac.uk; y.yang@ed.ac.uk). } 
\thanks{Ronald B. Liu is with the SMART Group at the Institute for Imaging, Data, and Communications, School of Engineering, The University of Edinburgh, EH9 3BF Edinburgh, U.K., and the Department of Biosystems, KU Leuven, 3001 Leuven, Belgium. (e-mail: ronald.liu@ed.ac.uk).}
\thanks{Francesco Giorgio-Serchi is with the Institute for Integrated Micro and Nano Systems, School of Engineering, The University of Edinburgh, EH8 9YL Edinburgh, U.K. (e-mail: F.Giorgio-Serchi@ed.ac.uk).}
\thanks{Peisan (Sharel) E is with the Institute for Bioengineering, School of Engineering, The University of Edinburgh, EH9 3DW Edinburgh, U.K. (e-mail: Sharel.E@ed.ac.uk).}
% \thanks{Huazhi Dong and Ronald B. Liu contributed equally to this work.}
}
\markboth{Journal of \LaTeX\ Class Files,~Vol.~14, No.~8, August~2021}%
{Shell \MakeLowercase{\textit{et al.}}: A Sample Article Using IEEEtran.cls for IEEE Journals}


\maketitle

\begin{abstract}
Electrical Impedance Tomography (EIT)-inspired tactile sensors are gaining attention in robotic tactile sensing due to their cost-effectiveness, safety, and scalability with sparse electrode configurations. This paper presents a data augmentation strategy for learning-based tactile reconstruction that amplifies the original single-frame signal measurement into 32 distinct, effective signal data for training. This approach supplements uncollected conditions of position information, resulting in more accurate and high-resolution tactile reconstructions. Data augmentation for EIT significantly reduces the required EIT measurements and achieves promising performance with even limited samples. Simulation results show that the proposed method improves the correlation coefficient by over 12\% and reduces the relative error by over 21\% under various noise levels. Furthermore, we demonstrate that a standard deep neural network (DNN) utilizing the proposed data augmentation reduces the required data down to 1/31 while achieving a similar tactile reconstruction quality. Real-world tests further validate the approach's effectiveness on a flexible EIT-based tactile sensor. These results could help address the challenge of training tactile sensing networks with limited available measurements, improving the accuracy and applicability of EIT-based tactile sensing systems.
\end{abstract}

\begin{IEEEkeywords}
Data augmentation, Electrical Impedance Tomography (EIT), robotic perception, tactile sensing, deep learning
\end{IEEEkeywords}

\section{Introduction}
% \label{sec:introduction}
\IEEEPARstart {R}{obotic} tactile perception is essential for natural and stable interactions, which significantly enhances the safety and intelligence of human-computer interactions \cite{pang2021review}. Inspired by the functionality of human skin, whole-body or large-scale tactile sensing requires sensors that are flexible, scalable, and capable of distributed sensing. These attributes allow the system to cover large areas without compromising mechanical compliance \cite{Park2021}. To fulfil these requirements, flexible tactile sensors consisting of discrete sensing elements \cite{SilveraTawil2015, Sundaram2019} or orthogonal stretchable electrodes \cite{Boutry2018, Won2019} are widely used. However, large-scale tactile sensing requires a considerable number of sensing elements or wires to maintain high spatial resolution, which limits its application.

Alternatively, recent advancements in tactile sensors \cite{sohn2017extremely, sakiyama2019evaluation, park2022biomimetic, luu2023simulation} have focused on inferring tactile stimuli by analyzing the distribution of tactile stimulus-response properties within a Region of Interest (ROI). In these designs, Electrical Impedance Tomography (EIT) has emerged as a promising method for whole-body/large-area tactile sensing owing to its sparse boundary electrode configurations and spatial resolving capabilities \cite{dong2024tactile, Tawil2011}. EIT reconstructs the conductivity distribution within the ROI by injecting a small current and measuring the induced voltage. Recent studies \cite{yoshimoto2024design,chen2023skin,liu2023robotic} have demonstrated the advantages of EIT-based tactile sensors, including safety, affordability, and ease of manufacturing, making EIT a viable solution for tactile perception in robotics.

Despite these advantages, the low spatial resolution of EIT, resulting from the inherently ill-posed and ill-conditioned inverse problem, continues to hinder its practical application in robotic perception \cite{adler2021electrical}. Therefore, there has been a surge of interest in leveraging end-to-end deep learning to enhance the performance of EIT-based tactile sensing \cite{husain2021tactile, chen2022convolutional, park2019deep}. Deep learning-based methods have shown promise in addressing the low spatial resolution challenge of EIT, enabling improved quality in EIT-based tactile reconstruction \cite{hardman2023tactile, chen2022large}. However, like all deep-learning methods, these models are heavily reliant on data quality, scale, and diversity. Collecting extensive and diverse tactile datasets from both simulations and real-world experiments presents a major bottleneck, significantly hindering effective model training with either simulated or real data.

Data augmentation is an effective technique to increase both the amount and diversity of data \cite{cubukAutoAugmentLearningAugmentation2019}; in computer vision algorithms, it has been proven to be effective in achieving state-of-the-art accuracy without additional data on standard datasets \cite{chenGridMaskDataAugmentation2024}, e.g., CIFAR10, CIFAR100 \cite{krizhevskyLearningMultipleLayers}, SVHN \cite{netzerReadingDigitsNatural}, and ImageNet \cite{dengImageNetLargescaleHierarchical2009}, etc. However, most studies on learning-based EIT tactile reconstruction have focused on neural network architectures (e.g., \cite{ma2022multi,husain2021tactile}), while less attention has been paid to data efficiency. Data pre-processing in EIT primarily focuses on adding noise to simulated data to improve reconstruction accuracy on real-world data \cite{Park2021}, or using fewer electrodes to simulate scenarios with a higher electrode count \cite{xu2022virtual}. They are insufficient as it is challenging to cover conditions of all positional information, leaving a gap in fully enhancing the data diversity and accuracy required for robust EIT tactile reconstruction.

Here, we introduce a data augmentation strategy for EIT-based tactile sensing. This approach amplifies a single-frame EIT measurement into 32 distinct EIT voltage signals, thereby significantly increasing the data scales without additional measurements. It is particularly effective in scenarios with limited measurements, enabling ``few-shot" measurements to generate an augmented, larger-scale dataset for training. Furthermore, the augmented data covers all possible conditions of effective measurements by rearranging the order of the original EIT measurement. As a result, the enhanced spatial resolution supplements the positional information omitted by the measurements. The resulting comprehensive dataset makes the training process based on the same measurement more robust. We validate the proposed approach through extensive simulations and real-world experiments, demonstrating its effectiveness in EIT-based tactile sensing, even with limited data for deep network training. The key contributions are as follows:
\begin{itemize}
\item We propose a data augmentation approach for the EIT-based tactile reconstruction, which converts a single EIT measurement into 32 distinct readouts, thereby enriching the dataset by supplementing uncollected location information.
\item We design a DNN model, a measurement vector is mapped into the latent variable through a multilayer perceptron (MLP), and a tactile map is generated from the latent variable through a convolutional neural network (CNN) for high-quality tactile reconstruction.
\item We develop a flexible EIT-based tactile sensor using carbon black and graphite, and validate the proposed data augmentation approach through real-world experiments.
\end{itemize}

%Our approach is validated through these results showing substantial improvements in tactile reconstruction performance with limited datasets, bridging the gap between deep learning and tactile sensing data acquisition challenges. Our strategy can be further adopted and adapted as a foundational paradigm for data preprocessing across all EIT-based sensing tasks. 
\section{Methodology}
\begin{figure}[t]
\centerline{\includegraphics[scale=0.85]{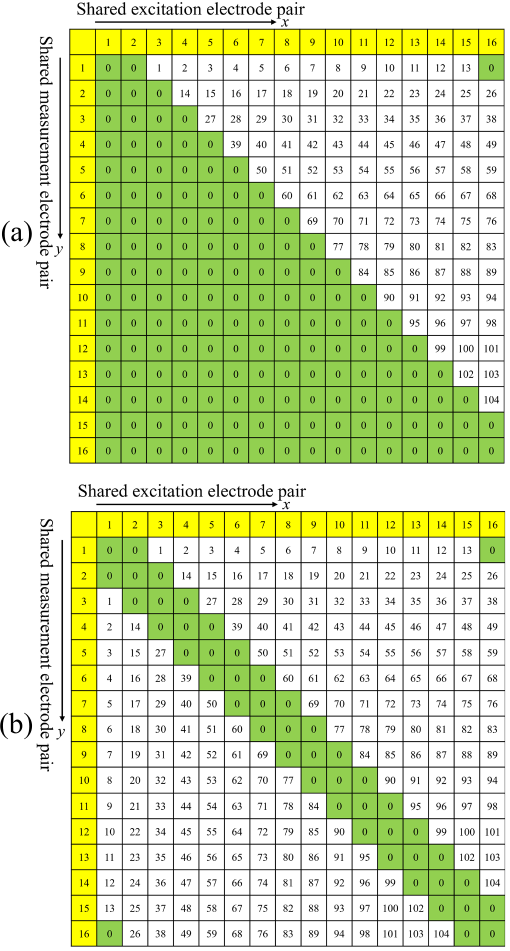}}
\caption{Electrical Impedance Maps (EIMs). (a) The EIM of 104 measurements. (b) The EIM of 208 measurements.}
\label{Fig1}
\end{figure}
\subsection{Principle of EIT-based Tactile Sensing}
The EIT-reconstruction problem in tactile sensing is to estimate the conductivity distribution $ \boldsymbol{\sigma} \in \mathbb{R}^n $ induced by touch in the ROI from the voltage measurements $ \mathbf{V} \in \mathbb{R}^m $. EIT-based tactile sensing usually adopts the time-difference imaging approach \cite{kuen2009multi}, which can be formulated as:
\begin{equation}
\Delta \mathbf{V}=\boldsymbol{J} \Delta \boldsymbol{\sigma}
\label{eq2}
\end{equation}
where $\boldsymbol{J} \in \mathbb{R}^{m \times n}$ is the Jacobian matrix, $\Delta \mathbf{V}$ is the voltage change, and $\Delta \boldsymbol{\sigma}$ is the conductivity change to a reference.
Under the model-based framework, the EIT inverse problem can generally be formulated as:
\begin{equation}
\Delta \hat{\sigma} =\arg \min _{\Delta \boldsymbol{\sigma}}\|\boldsymbol{J} \Delta \boldsymbol{\sigma}-\Delta \mathbf{V}\|_2^2+\tau R(\Delta \boldsymbol{\sigma})
\label{eq3}
\end{equation}
where $R$ represents the regularizer that incorporates prior information and $\tau>0$ is the regularisation factor. Alternatively, the learning-based framework aims to find an inverse mapping operator $F^{-1}$ via data-driven approaches:
\begin{equation}
\Delta \hat{\sigma}=F^{-1}(\Delta V)
\label{eq6}
\end{equation}

\subsection{Electrical Impedance Map (EIM)}
\begin{figure*}[t]
\centerline{\includegraphics[width=\textwidth]{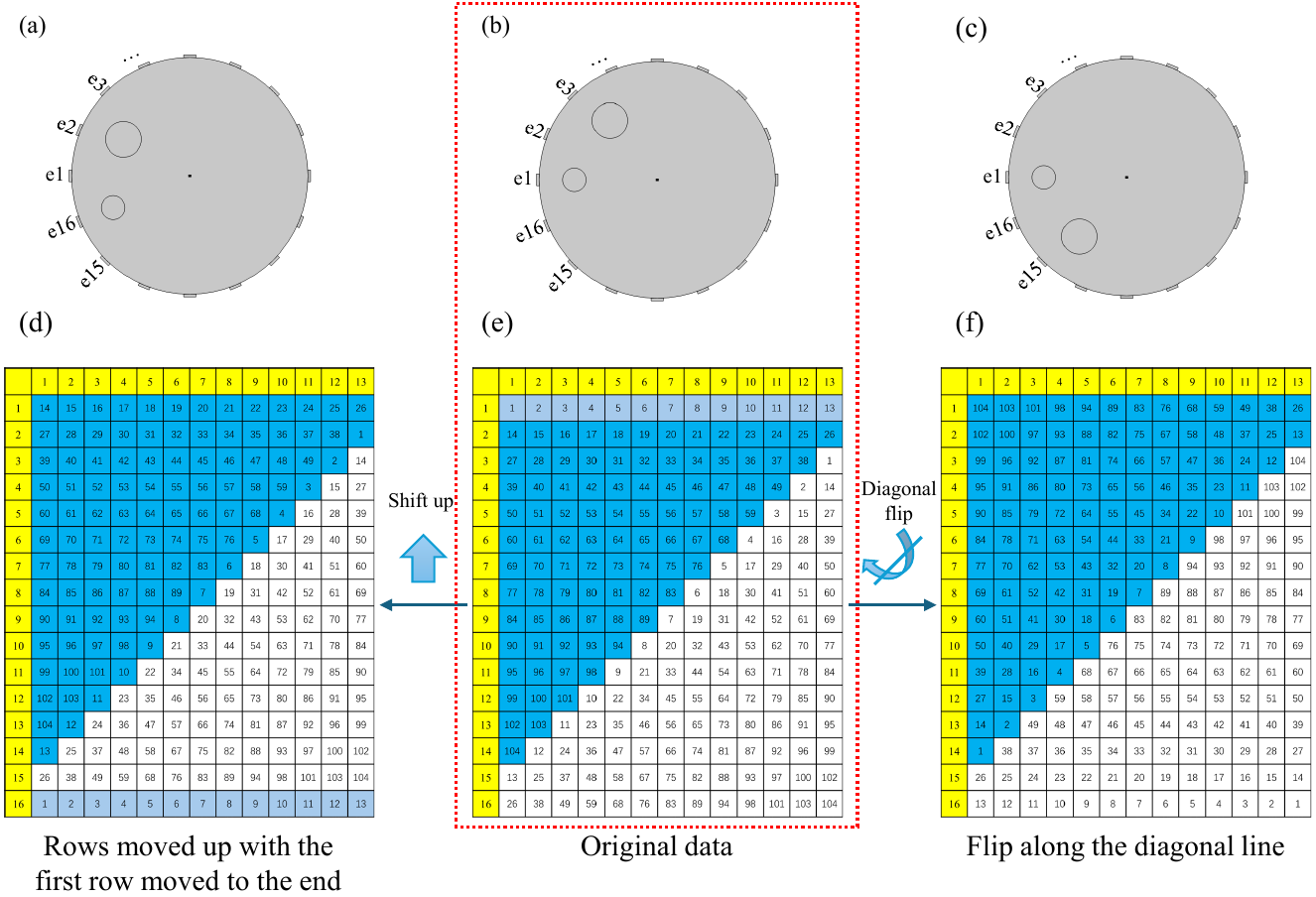}}
\caption{Illustration of the proposed data augmentation strategy. (a) - (c) Different positions of touch represented through rotational and flip transformations, demonstrating the rotation and flip relationship between (a), (b), and (c). (d) - (f) The corresponding EIM of different positions touch. The blue part shows 104 original EIT measurements. (d) is derived by shifting the rows of (e) upwards, and (f) is derived by flipping (e) along the diagonal.}
\label{Fig3}
\end{figure*}
\begin{figure*}[t]
\centerline{\includegraphics[scale=1]{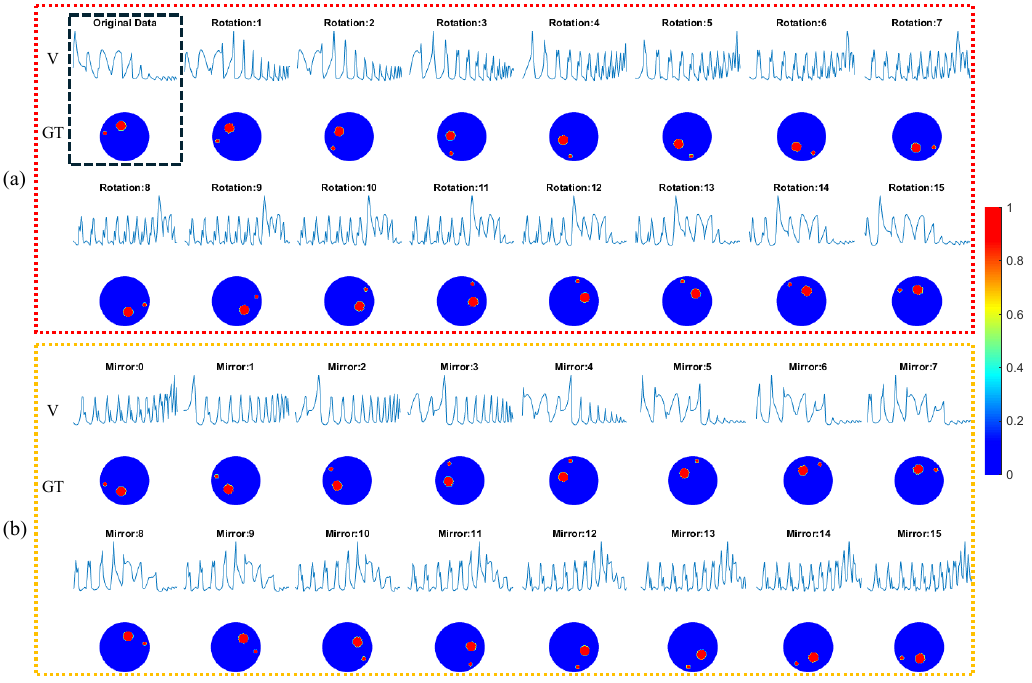}}
\caption{Two examples using the proposed data augmentation approach. (a) Rotation transformations are applied to the original data. (b) Flip transformations are applied to (a).}
\label{Fig4}
\end{figure*}
The Electrical Impedance Map (EIM) matrix provides a transformed representation of raw EIT measurements \cite{hu2019image}. Converting the data from a sequence to a matrix form (e.g., the transformation from (104,1) to (16,16), see Fig. \ref{Fig1}a) not only captures the geometric characteristics of EIT sensors but also facilitates data augmentation through spatial transformations. In a single EIT measurement, two neighbouring electrodes are used for current injection, while the remaining electrodes are selected for voltage measurements. Among the measuring electrodes, measurements are taken between the neighbouring electrode pairs, excluding the current injection ones. For our tactile sensor with 16 electrodes, 13 measurements are recorded following each excitation, resulting in a total of 16 (excitation electrode pairs) × 13 (measurement electrode pairs) = 208 original measurements. According to the reciprocity theorem, there are 104 non-redundant measurements in a sequential format in each frame.

One EIM matrix for a sequential measurement mentioned above is constructed as follows: As shown in Fig. \ref{Fig1}a, the differential voltage measurement between the $i_{th}$ and  ${(i+1)}_{th}$ electrodes, denoted as $V_{(i,i+1)}^{(j,j+1)}$, where $e_{i}$ and $e_{i+1}$ are the measurement electrodes and $e_{j}$ and $e_{j+1}$ are the excitation electrodes, is placed at the $(i,j)$ position in the EIM matrix, with remaining elements padded with zeros (see Fig. \ref{Fig1}a).  Each row in the matrix corresponds to a unique excitation electrode pair, and each column corresponds to a measurement electrode pair. According to the reciprocity theorem, e.g., $V_{(3,4)}^{(1,2)}$ is equal to $V_{(1,2)}^{(3,4)}$, the original 104 measurements can be extended to 208 elements, which are arranged into the 16 $\times$ 16 EIM matrix, as shown in Fig. \ref{Fig1}b.
\newcommand{\Input}{\item[\textbf{Input:}]}
\newcommand{\Output}{\item[\textbf{Output:}]}
\begin{algorithm}
\caption{Data Augmentation for 16 different positions.}
\label{alg1}
\begin{algorithmic}[1]
\Input $EIM$: Matrix of size 16 $\times$ 13 representing original voltage data
\Output $Index\_V16$: A 16 $\times$ 104 matrix
\State Initialize $Index\_V16$ as a 16 $\times$ 104 zero matrix
\For{$time = 1$ to 16}
    \State Set $index$ to 1
    \State Set $ele\_index$ to $time$
    \For{$i = 1$ to 14}
        \If{$i == 1$}
            \State Set $num$ to 13
        \Else
            \State Set $num$ to $15 - i$
        \EndIf
        \State Set $Index\_V16(time, index:index+num-1)$ to $EIM(ele\_index, 1:num)$
        \State Set $index$ to $index + num$
        \State Set $ele\_index$ to $ele\_index + 1$
        \If{$ele\_index == 17$}
            \State Set $ele\_index$ to 1
        \EndIf
    \EndFor
\EndFor
\State \Return $Index\_V16$
\end{algorithmic}
\end{algorithm}
\begin{algorithm}
\caption{Data Augmentation for 16 mirror positions.}
\label{alg2}
\begin{algorithmic}[1]
\Input $EIM$: Matrix of size 16 $\times$ 13 representing original voltage data
\Output $Index\_V16\_inv$: A 16 $\times$ 104 matrix
\State Initialize $Index\_V16\_inv$ as a 16 $\times$ 104 zero matrix
\State Flip each row of $EIM$ to get $EIM\_flipped$ by reversing the order of columns
\For{$time = 1$ to 16}
    \State Set $index$ to 1
    \State Set $ele\_index$ to $time$
    \For{$i = 1$ to 14}
        \If{$i == 1$}
            \State Set $num$ to 13
        \Else
            \State Set $num$ to $15 - i$
        \EndIf
        \State Set $Index\_V16\_inv(time, index:index+num-1)$ to $EIM\_flipped(ele\_index, 1:num)$
        \State Set $index$ to $index + num$
        \State Set $ele\_index$ to $ele\_index + 1$
        \If{$ele\_index == 17$}
            \State Set $ele\_index$ to 1
        \EndIf
    \EndFor
\EndFor
\State \Return $Index\_V16\_inv$
\end{algorithmic}
\end{algorithm}

We then extract the 13 valid values from each row in Fig. \ref{Fig1}b and sort them according to the EIT measurement strategy to create a 16 $\times$ 13 EIM matrix, e.g., see Fig. \ref{Fig3}e.

\subsection{Data Augmentation Strategies}
In this work, we focus on circular tactile sensors where EIT electrodes are evenly distributed around the circumference. This symmetrical arrangement enables us to apply effective data augmentation strategies to enhance the robustness and diversity of our dataset. Although our current approach is tailored to circular sensors, the underlying principles of these data augmentation strategies can be adapted to other sensor shapes. For non-circular geometries, similar augmentation techniques could be developed by exploiting the specific symmetries or repeating patterns in the electrode arrangements, allowing for a broader application of these methods across various sensor designs. As shown in Fig. \ref{Fig3}, we propose two strategies of data augmentation based on EIM.

\textbf{Strategy 1: Rotating} - Given the circular symmetry of the tactile sensors, we can leverage the \SI{22.5}{\degree} rotational intervals between electrodes to generate new data frames from existing ones. Specifically, in Fig. \ref{Fig3}, the difference between Fig. \ref{Fig3}a and Fig. \ref{Fig3}b lies in the position of the touch. The touch positions in Fig. \ref{Fig3}a correspond to a \SI{22.5}{\degree} rotation of the touch in Fig. \ref{Fig3}b. In Fig. \ref{Fig3}a, $e_{1}$ and $e_{2}$ are selected as the excitation electrode pairs, the differential voltage measurements $V_{(3,4)}^{(1,2)}$ for $e_{3}$ and $e_{4}$ should be the same as the measurement $V_{(4,5)}^{(2,3)}$ for $e_{2}$ and $e_{3}$ when $e_{16}$ and $e_{1}$ are excitements in Fig. \ref{Fig3}b. Therefore, we can generate the EIM matrix (Fig. \ref{Fig3}d) for the corresponding Fig. \ref{Fig3}a using this voltage equivalence method. 

Following the Algorithm \ref{alg1}, a single frame of EIT data can be expanded into 16 frames (see Fig. \ref{Fig4}a), effectively increasing the dataset scale 16 times without requiring additional physical measurements. A set of rotations can capture the positional information across all degrees, thereby enhancing the model's ability to generalize across different touch positions and improving the robustness of the tactile reconstruction.

\textbf{Strategy 2: Flipping} - The tactile position in Fig. \ref{Fig3}c is the mirror position of Fig. \ref{Fig3}b across the horizontal axis. In Fig. \ref{Fig3}c, when we select $e_{1}$ and $e_{2}$ as the excitation electrode pairs, the differential voltage measurements $V_{(3,4)}^{(1,2)}$ for $e_{3}$ and $e_{4}$ should be the same as the  $V_{(15,14)}^{(1,16)}$ for $e_{15}$ and $e_{14}$ when $e_{1}$ and $e_{16}$ are the excitation electrode pairs in Fig. \ref{Fig3}b. We can generate the corresponding EIM matrix (Fig. \ref{Fig3}f) for Fig. \ref{Fig3}c using the above voltage equivalence method. 

Following the strategy in Algorithm \ref{alg2}, the data in Fig. \ref{Fig4}a can be further expanded to 16 additional frames of data, as shown in Fig. \ref{Fig4}b. The result in Fig. \ref{Fig4}b is a mirror inversion of the position corresponding to Fig. \ref{Fig4}a. This flipping strategy not only increases the dataset size further but also introduces variations in the sensor's response to mirrored touch positions, aiding in learning more complex spatial relationships.

Note that the generated EIM for these two strategies is only used during data augmentation, while we recover the original EIT measurement format from the EIM during tactile reconstruction.

\section{Numerical Simulation}
\subsection{Dataset and Network Training}
The proposed data augmentation approach was validated on the Edinburgh EIT Dataset (EdEIT) \cite{Chen2022}, which was generated via COMSOL Multiphysics. In this dataset, the conductivity for the background media is set as 0.05\,S/m, and all objects' conductivity varies between 0.0001 and 0.05\,S/m. The dataset emphasizes position and tactile intensity, with all objects being circular and having random diameters. EdEIT comprises 29,333 samples, including 7,035 samples with 1 circle, 7,298 samples with 2 circles, 7,500 samples with 3 circles, and 7,500 samples with 4 circles. Each sample consists of a 1 $\times$ 104-dimensional vector for the voltage measurement and a 64 $\times$ 64-dimensional matrix for the corresponding conductivity distribution.

We split the dataset into three subsets: a training set containing 21,833 samples with 1, 2, and 3 touching objects; a validation set with 5,000 samples featuring 4 touching objects; and a test set with 2,500 samples containing 4 touching objects. To enhance the model generalisation ability, additive white Gaussian noise with a Signal-to-Noise Ratio (SNR) of 50dB was added to the training data for data augmentation. 

\begin{figure*}[t]
\centerline{\includegraphics[scale=0.55]{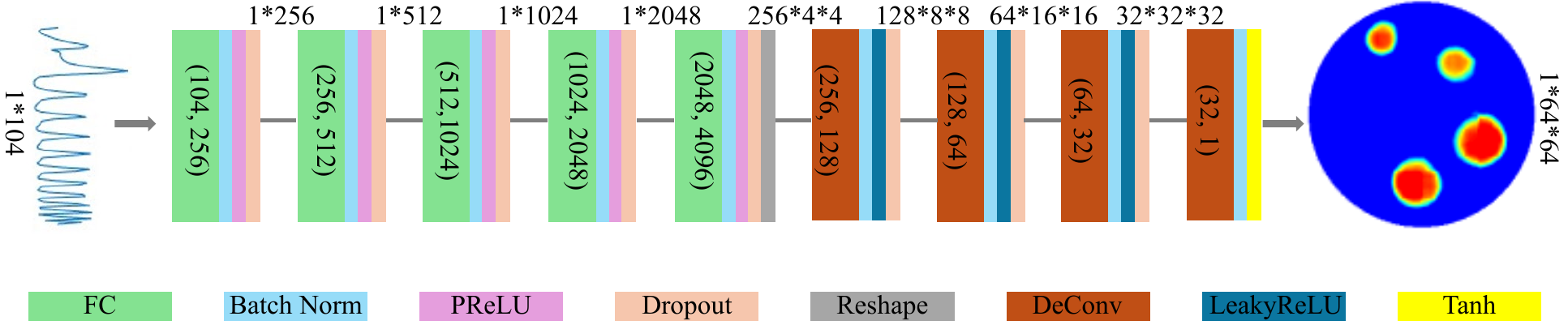}}
\caption{The architecture of the DNN model for tactile reconstruction.}
\label{Fig5}
\end{figure*}
\begin{figure}[t]
\centerline{\includegraphics[scale=0.45]{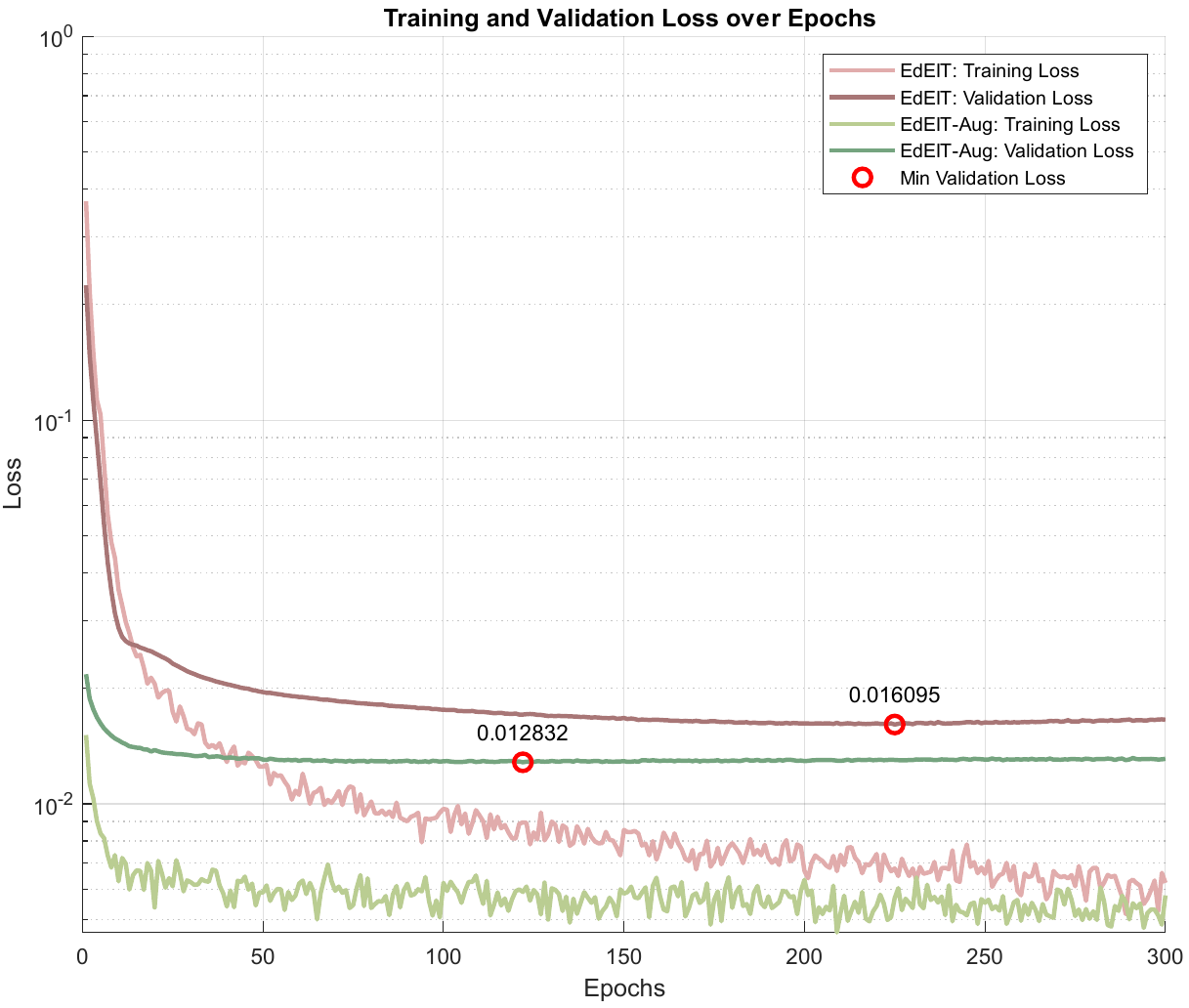}}
\caption{Training and validation loss.}
\label{Fig6}
\end{figure}
The study is evaluated on a standard deep neural network (DNN) built with multilayer perceptron (MLP) and convolutional neural network (CNN), a framework commonly used in related works \cite{Park2021,park2022biomimetic, chen2022convolutional}. As shown in Fig. \ref{Fig5}, the input is the EIT measurement with a sequence of voltages, while the output is an image representing the conductivity within the ROI. In this method, the input measurement vector is encoded to the latent variables via the MLPs and the conductivity distribution is decoded from the latent variables via the CNNs.

For training, we utilized the Mean Squared Error (MSE) loss function and employed the Adam optimizer \cite{Kingma2014} with an initial learning rate of 0.0001 for 300 epochs, with a batch size of 512. To accelerate the training process and improve model performance, the DNN model adopts a Batch Normalisation strategy \cite{Ioffe2015}. In addition, by introducing the dropout strategy, the model effectively alleviates the over-fitting problem and enhances its generalisation ability on unseen data. Fig. \ref{Fig6} showcases the training and validation results in MSE loss; the augmented EdEIT (EdEIT-Aug) data through our data augmentation strategy yields superior results compared to the original EdEIT datasets. Data augmentation accelerates model convergence and reduces the validation MSE loss from 0.0161 to 0.0128.  All training was conducted on a computer equipped with an Intel i9-13900HX CPU and an RTX 4080 GPU.

\subsection{Results and Discussion}
\begin{figure*}[t]
\centerline{\includegraphics[scale=0.75]{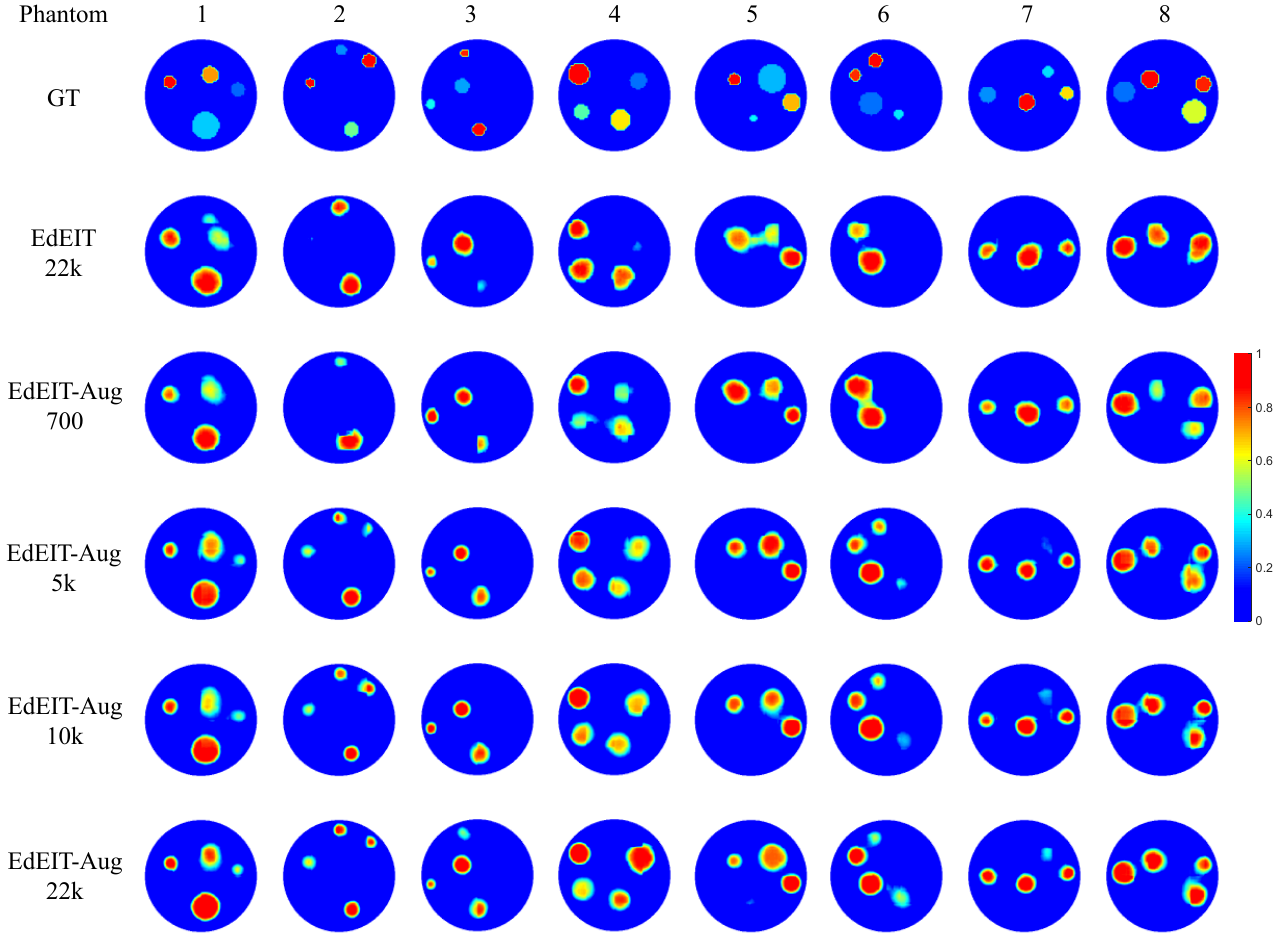}}
\caption{Tactile reconstruction based on simulation data (SNR = 50 dB). All results are normalized .}
\label{Fig7}
\end{figure*}
\begin{figure}[t]
\centerline{\includegraphics[scale=0.3]{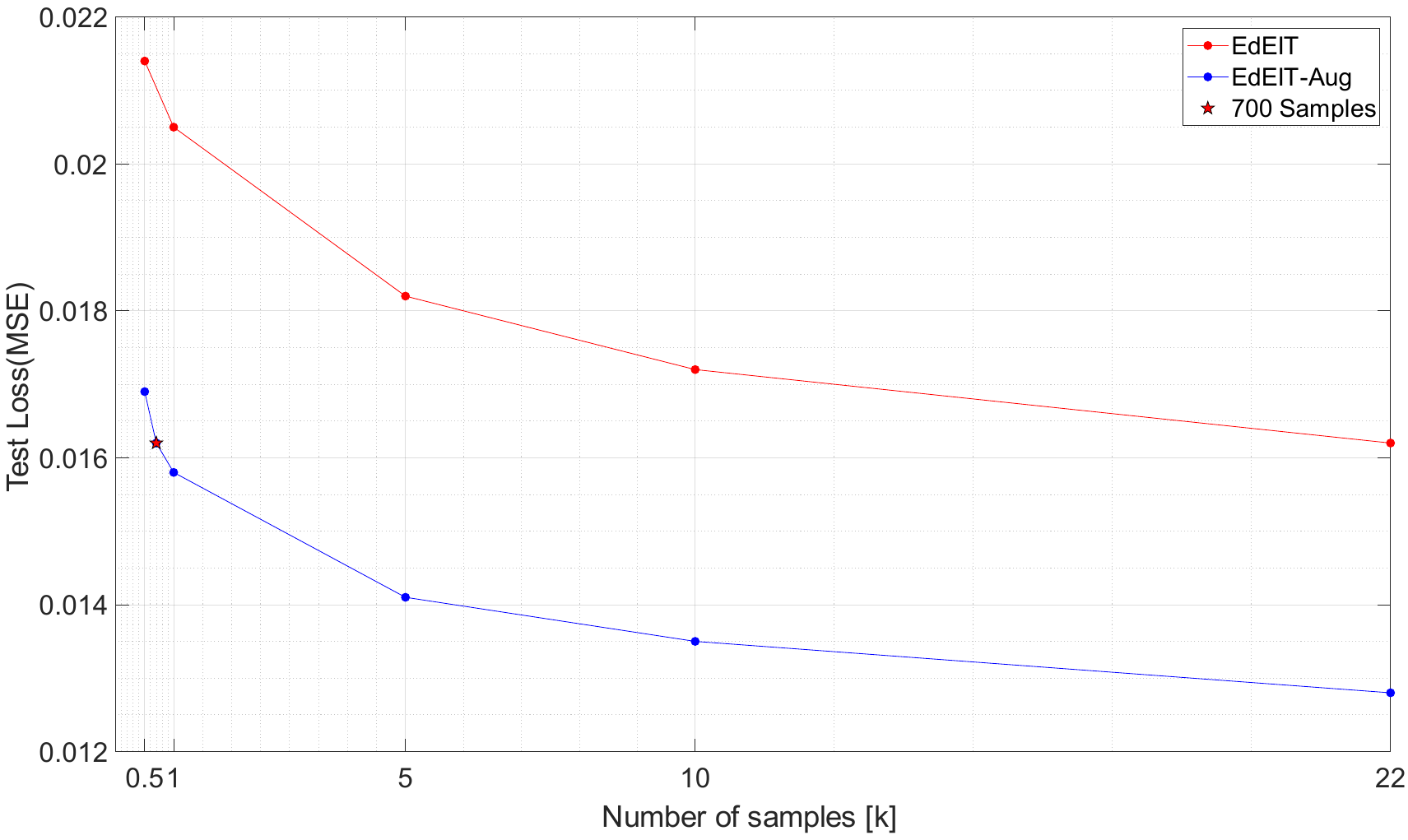}}
\caption{The test loss of different numbers of training data.}
\label{Fig8}
\end{figure}
We employ the Correlation Coefficient (CC) and Relative Error (RE) \cite{dong2024tactile} to assess tactile reconstruction performance, averaging results over 2500 test samples. These quantitative metrics (see Table \ref{tab1}) demonstrate the effectiveness of the proposed data augmentation method on model performance. Among all datasets, the DNN model using the EdEIT-Aug achieves the highest CC and the lowest RE values across all levels of noise input, indicating a superior performance in reconstructing fine details of tactile interactions. Notably, the CC values for the EdEIT-Aug show an improvement of approximately 12.82\% to 17.35\% compared to the standard EdEIT, with the most significant improvement observed in the noise-free condition (from 0.5903 to 0.6927). Similarly, the RE values for the EdEIT-Aug demonstrate a reduction of approximately 21.19\% to 23.74\%, with the most notable reduction also in the noise-free condition (from 1.1030 to 0.8412). This underscores the critical role of the proposed data augmentation approach in improving the quality of tactile reconstructions.
\begin{table}[t]
\centering
\caption{Quantitative metrics.}
\label{tab1}
\scriptsize % 或 \footnotesize
\begin{tabular}{|c|c|c|c|c|c|}
\hline
Metrics              & SNR(dB)    & EdEIT    & EdEIT-Aug           & \begin{tabular}[c]{@{}c@{}}CC \\ Improvement\end{tabular} & \begin{tabular}[c]{@{}c@{}}RE \\ Reduction\end{tabular} \\ \hline
\multirow{4}{*}{CC}  & 30         & 0.5863 & \textbf{0.6681} & 13.95\%                                                   & -                                                        \\ \cline{2-6} 
                     & 40         & 0.5899 & \textbf{0.6662} & 12.93\%                                                   & -                                                        \\ \cline{2-6} 
                     & 50         & 0.5903 & \textbf{0.6660} & 12.82\%                                                   & -                                                        \\ \cline{2-6} 
                     & Noise Free & 0.5903 & \textbf{0.6927} & 17.35\%                                                   & -                                                        \\ \hline
\multirow{4}{*}{RE} & 30         & 1.1071 & \textbf{0.8667} & -                                                         & 21.71\%                                                  \\ \cline{2-6} 
                     & 40         & 1.1033 & \textbf{0.8689} & -                                                         & 21.25\%                                                  \\ \cline{2-6} 
                     & 50         & 1.1029 & \textbf{0.8692} & -                                                         & 21.19\%                                                  \\ \cline{2-6} 
                     & Noise Free & 1.1030 & \textbf{0.8412} & -                                                         & 23.74\%                                                  \\ \hline
\end{tabular}
\vspace{0.2mm}
\begin{minipage}{\columnwidth}
\raggedright
\footnotesize{ The best results are highlighted in bold.}
\end{minipage}
\end{table}

Additionally, we randomly selected 8 samples from the 2500 test set reconstructions to visually demonstrate the performance differences. As shown in Fig. \ref{Fig7}, the Ground Truth (GT) images in the first row serve as references.  The subsequent rows present reconstructions using the EdEIT and EdEIT-Aug, respectively. From these visual results, it is evident that the EdEIT-Aug, which incorporates data augmentation, provides significantly higher accuracy in reconstructions than the standard EdEIT dataset. Specifically, the model with EdEIT-Aug reconstructs the number of touch objects more accurately and reveals sharper object edges. 

To further illustrate the effectiveness of data augmentation, we compared the performance of the model trained with different amounts of training data from EdEIT and EdEIT-Aug, respectively. Fig. \ref{Fig8} shows the test loss (MSE) for models trained on varying sample sizes, from 700 to 21,833 samples.

The results indicate that applying data augmentation significantly improves the model's performance, even with a small training set. Notably, the model trained with EdEIT-Aug augmented from 700 samples achieves a comparable test loss to the one trained on the full training dataset of 21,833 samples from EdEIT. This demonstrates our data augmentation strategy can effectively reduce the required raw EIT signal measurements by over 31 times while maintaining the same quality of reconstruction. Such a reduction in measurement requirement is particularly valuable in scenarios with limited sample availability, highlighting the potential of data augmentation to make efficient use of limited data and achieve high performance in tactile reconstruction tasks.

\section{Real-world Experiments}
\subsection{Sensor Fabrication}
\begin{figure}[t]
\centerline{\includegraphics[scale=0.65]{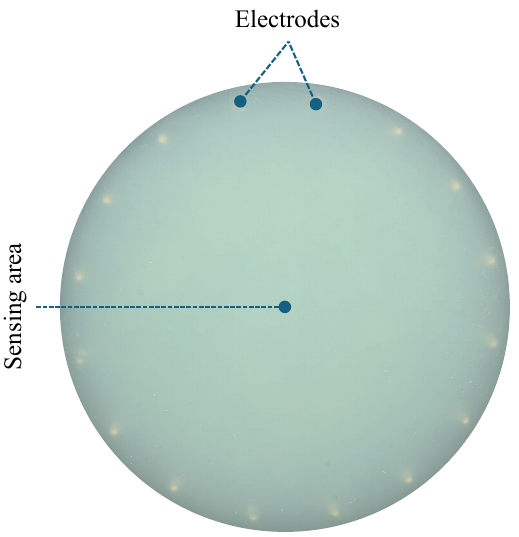}}
\caption{The fabricated EIT-based tactile sensor.}
\label{fig-sensorfab}
\end{figure}
Fig. \mbox{\ref{fig-sensorfab}} illustrates the EIT-based tactile sensor, which comprises a substrate layer fabricated by silicone, a sensing layer fabricated by hydrogel \mbox{\cite{dong2024tactile}}, a sealing layer fabricated by silicone, and 16 evenly distributed boundary electrodes. The sensor has an overall circular shape with a diameter of 120 mm and a thickness of 100 mm. The conductive layer inside the sensor maintains the same circular shape with a diameter of 120 mm and a thickness of 40 mm. The substrate and sealing layers have a diameter of 120 mm and a thickness of 30 mm. Each electrode forms a circular contact area with a diameter of 3 mm with the conductive layer to ensure effective signal transmission.

\subsection{Results and Discussion}
\begin{figure*}[t]
\centering
\includegraphics[width=\textwidth]{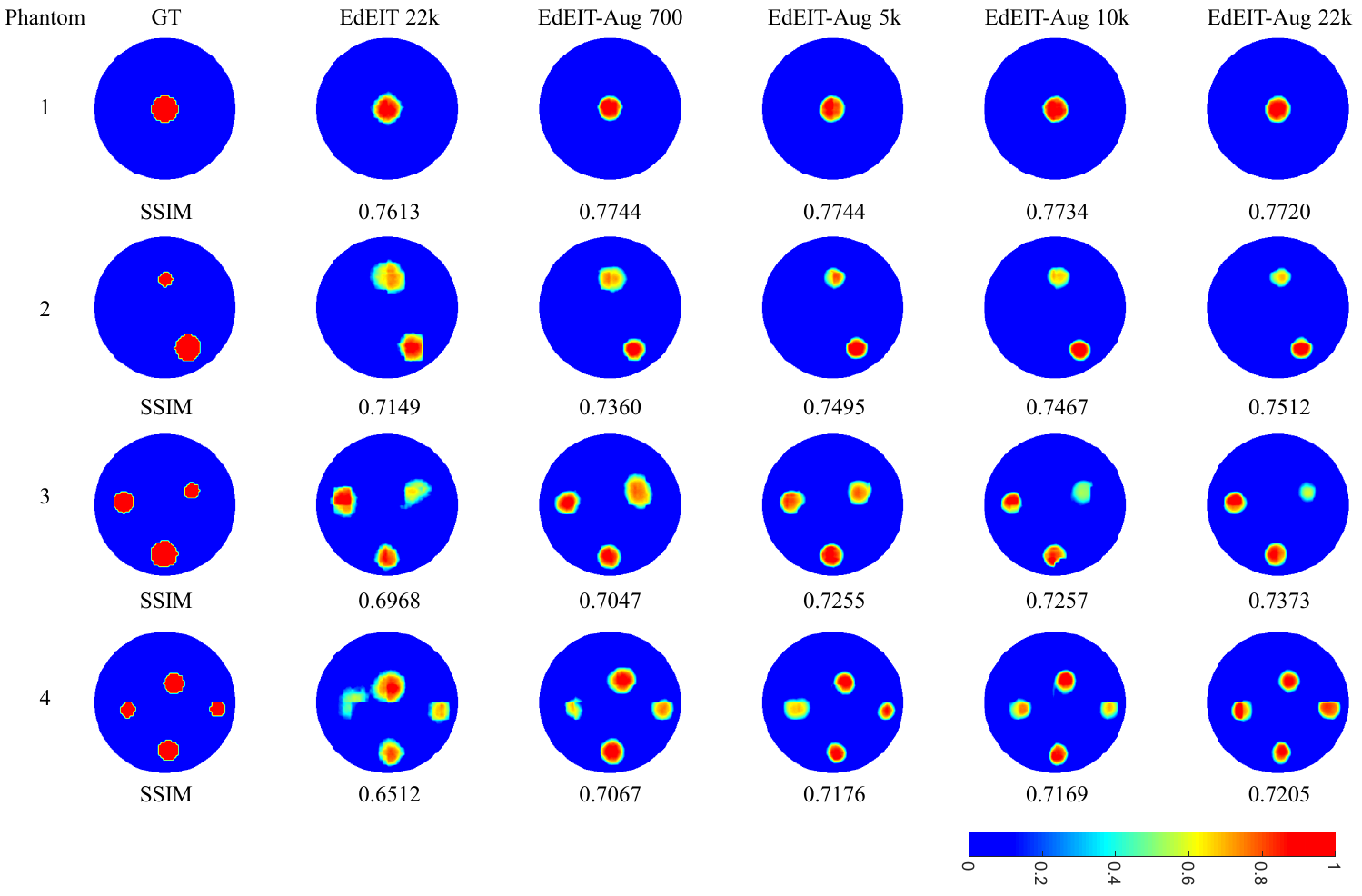}
\caption{Tactile reconstructions based on real-world experiment data. All results are normalized.}
\label{fig-experimentresult1}
\end{figure*}
\begin{figure*}[t]
\centering
\includegraphics[width=\textwidth]{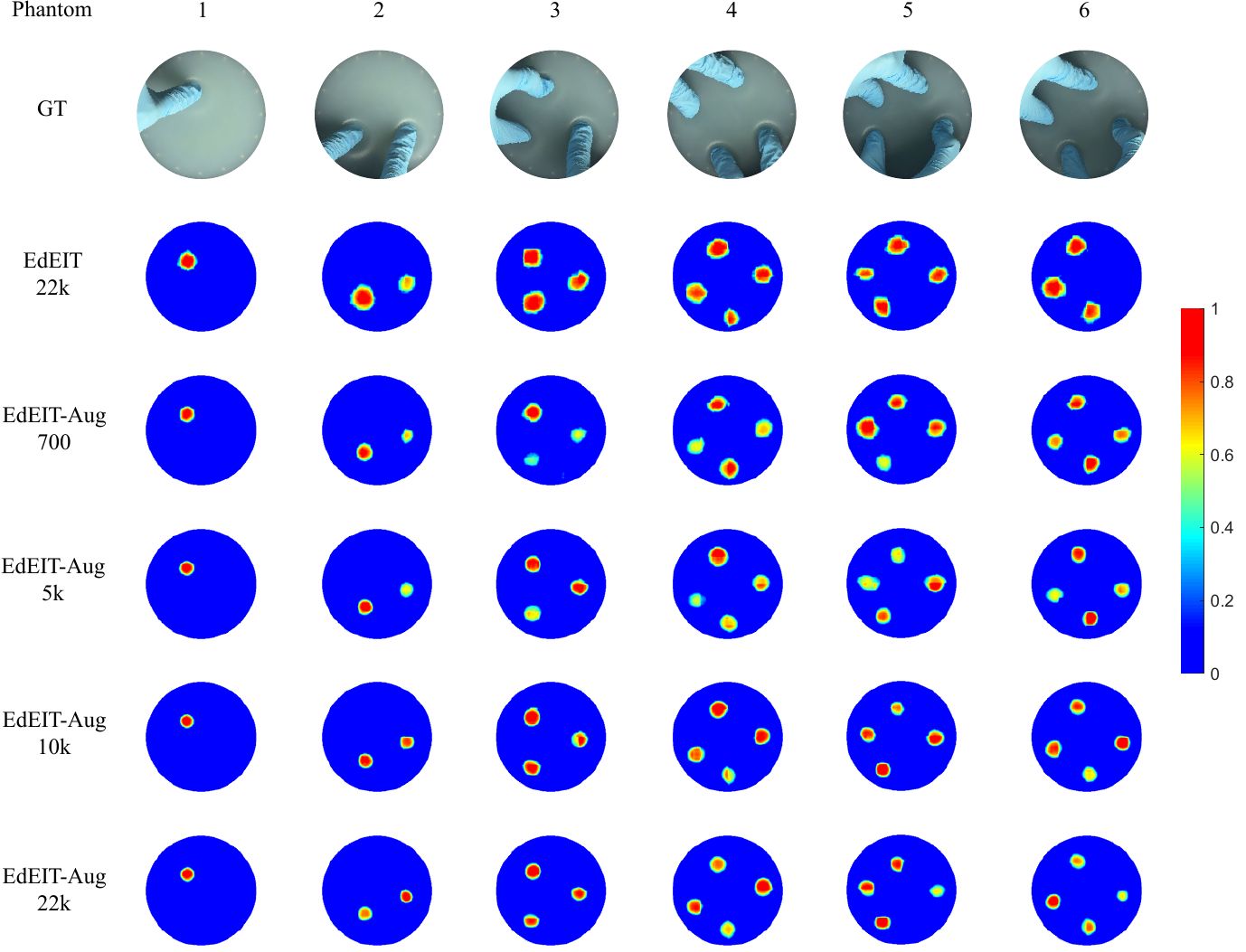}
\caption{Tactile reconstruction for complex finger touches. All results are normalized.}
\label{fig-experimentresult2}
\end{figure*}
We conducted real-world experiments to evaluate the effectiveness of the proposed method. To this end, we directly deployed the DNN model trained using simulation data onto real-world tactile measurements. A multi-frequency EIT system was used for data acquisition, with adjacency excitation and adjacency measurement scheme \cite{Yang2017}.

As shown in Fig. \ref{fig-experimentresult1}, from Phantom 1 to Phantom 4, we applied 1 to 4 contacts in different positions on the sensor surface using the room linear robot (DLE-RG-0001). Similar to the simulation, we visualized these phantoms of the reconstruction.  We employ the Structural Similarity (SSIM) \cite{wang2004image} to assess tactile reconstruction performance. Note that SSIM is particularly suitable in this context because we only can get the positions and sizes of the touch points through the room linear robot, making it an effective quantitive metric for capturing the structural similarities of the tactile images. The results clearly show that, for all phantoms, the dataset after 700 frames of data augmentation  (EdEIT-Aug 700) has reached the results on EdEIT using all data (21,833 frames). As the amount of enhanced data included increases, the tactile reconstruction becomes more accurate, and the touch edges become sharper. This demonstrates that the data augmentation method is equally effective on real-world experimental data.

In addition to the above experiments, we further validate the effectiveness of the proposed method using real-world tactile measurements with more complex and natural touch patterns. These include configurations involving one to four finger touches, which generate irregular pressure profiles. The tactile reconstruction results are shown in Fig. \ref{fig-experimentresult2}, the proposed data augmentation approach enables the DNN model to capture the number of reconstructed touch points and the fine details of the pressure variations accurately. Notably, for Phantom 6, the DNN model without data augmentation fails to reconstruct the correct number of touch points, whereas even the DNN model trained with only 700 augmented frames can accurately capture the number of touches. These results not only demonstrate the robustness of our method in handling intricate real-world interactions but also highlight its potential for practical applications in advanced tactile sensing systems.

\section{Conclusion}
This paper presents a data augmentation strategy for EIT-based tactile sensing that effectively addresses the challenge of limited datasets by amplifying a single-frame signal into 32 distinct signals. This data augmentation approach significantly improves tactile reconstruction accuracy with the same DNN model, with correlation coefficients increasing by up to 17.35\% and relative errors decreasing by up to 23.74\%. Moreover, it dramatically reduces the need for measurement to build large training datasets, achieving comparable performance with over 31 times less raw data. The augmented dataset delivers superior reconstruction, accurately capturing touch details and providing clearer edges. This method is particularly effective in scenarios where data collection is limited, bridging the gap between deep learning and tactile sensing data acquisition challenges. Our strategy could serve as a foundational paradigm for data preprocessing across all EIT-based sensing tasks.

\bibliographystyle{IEEEtran}
\bibliography{References}

\vfill
\end{document}